\pdfoutput=1

\documentclass[11pt]{article}

\usepackage[preprint]{acl}

\usepackage{times}
\usepackage{latexsym}
\usepackage{fontawesome5}

\makeatletter
\newcommand{\github}[1]{%
   \href{#1}{\faGithubSquare}%
}
\makeatother

\usepackage[T1]{fontenc}

\usepackage[utf8]{inputenc}

\usepackage{microtype}

\usepackage{inconsolata}

\usepackage{graphicx}

\usepackage{hyperref}       
\usepackage{url}            
\usepackage{booktabs}       
\usepackage{amsfonts}       
\usepackage{nicefrac}       
\usepackage{xcolor}         

\usepackage{amsmath}
\usepackage{array}
\usepackage{colortbl}
\usepackage{wrapfig}
\usepackage{cleveref}
\usepackage{todonotes}
\usepackage{enumitem,kantlipsum}
\usepackage{tikz}
\usetikzlibrary{arrows.meta, positioning, shapes.geometric, shadows}

\newcommand{\ZSEVAL}{\textsc{ZeroSumEval}}

\definecolor{zse_color1}{RGB}{41,44,147}
\definecolor{zse_color2}{RGB}{14,140,247}

\title{%
  \begin{tabular}{cc}
    \raisebox{-0.35\height}{\includegraphics[height=45pt]{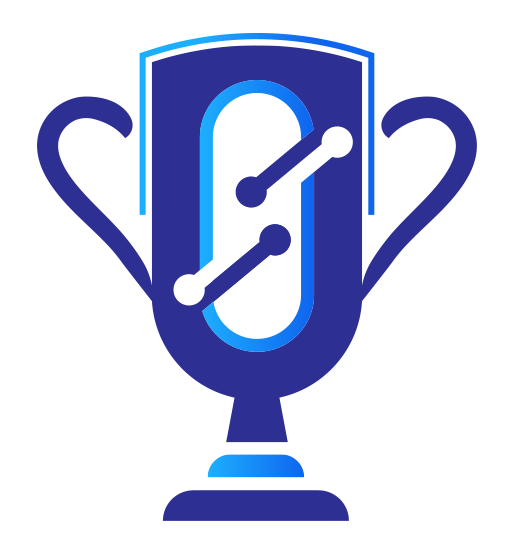}} &
    \begin{tabular}{c}
      \ZSEVAL{}: An Extensible Framework For Scaling \\
      LLM Evaluation with Inter-Model Competition
    \end{tabular}
  \end{tabular}%
}

\author{
 \textbf{Hisham A. Alyahya\textsuperscript{1,*}},
 \textbf{Haidar Khan\textsuperscript{2,*}},
 \textbf{Yazeed Alnumay\textsuperscript{3}},
 \\
 \textbf{M Saiful Bari\textsuperscript{1}},
 \textbf{Bülent Yener\textsuperscript{4}}
\\
 \textsuperscript{1}Saudi Data \& AI Authority (SDAIA),
 \textsuperscript{2}Meta,
 \textsuperscript{3}Cohere,
 \textsuperscript{4}Rensselaer Polytechnic Institute
\\
*Core contributors
 \\
    \textbf{\href{https://github.com/facebookresearch/ZeroSumEval}{\faGithub \hspace{0.05cm} https://github.com/facebookresearch/ZeroSumEval}}
 \\
 \small{
    Correspondence to: \href{mailto:haidark@meta.com}{haidark@meta.com}
 }
}

\begin{document}
\maketitle
\begin{abstract}
We introduce \ZSEVAL{}, a dynamic, competition-based, and evolving evaluation framework for Large Language Models (LLMs) that leverages competitive games. \ZSEVAL{} encompasses a diverse suite of games, including security challenges (Capture the Flag), classic board games (chess), and knowledge tests (MathQuiz). These games are designed to evaluate a range of capabilities such as strategic reasoning, planning, knowledge application, safety, and adaptability. Building upon recent studies that highlight the effectiveness of game-based evaluations for LLMs, \ZSEVAL{} enhances these approaches by providing a standardized and extensible framework for easily implementing games and leverages DSPy to provide a better abstraction for LLM player strategies.

\end{abstract}

\section{Introduction}
\label{sec:introduction}

Evaluation and benchmarking of Large Language Models (LLMs) is largely done in a \textit{static} manner, by building a test set for a particular task and running models against it and checking whether the model output matches what is expected. This direction suffers from multiple weaknesses: \textit{(i)} Data contamination \citep{yang2023rethinking}, where models inadvertently train on portions of the test data \citep{llama3,groeneveld2024olmoacceleratingsciencelanguage}, leading to inflated performance metrics. \textit{(ii)} Sensitivity to prompt variations \citep{alzahrani2024benchmarks} and a lack of diversity in evaluation tasks \citep{laskar2024systematic} further undermine the reliability and robustness of these benchmarks. \textit{(iii)} A high cost and effort required to develop new benchmarks often result in outdated evaluation methods that do not keep pace with the rapid development of LLMs \citep{kiela2021dynabenchrethinkingbenchmarkingnlp,vu2023freshllms,phan2025humanitysexam}.

Recent research has attempted to address the limitations of static LLM evaluation by introducing \textit{dynamic} evaluation methods that more effectively assess model performance \citep{zhuge2024agentasajudgeevaluateagentsagents,xu2024crabcrossenvironmentagentbenchmark,fan2024nphardevaldynamicbenchmarkreasoning,yu2024kievalknowledgegroundedinteractiveevaluation,liu2024agentbench,zhou2023webarena}. These approaches move beyond traditional static evaluation methodologies by creating dynamic environments in which LLMs are evaluated. This has demonstrated greater robustness in benchmarking LLM capabilities (further discussed in Section \ref{sec:related_works}).

This is certainly a step in the right direction; our work continues in this direction by posing evaluation strictly as competition between models. As models rapidly improve, they continually push against and even surpass existing benchmarks, leading to score saturation and diminishing the benchmarks’ usefulness. Furthermore, in most real-world scenarios, the primary goal of evaluation is not to determine how well a model performs in isolation, but rather to compare models relative to each other. This makes ranking more important than raw scores. We propose that competition between models in simulated game environments is an evaluation protocol that addresses these needs. By pitting models against other models, we ensure that models are compared directly against each other, and not against predetermined definitions of performance. This results in an evaluation protocol that is scalable; evolving alongside model capabilities to make tasks harder as models improve.

Previous work has also proposed the use of games as benchmarks \citep{topsakal2024evaluatinglargelanguagemodels}, offering a promising avenue for evaluating complex reasoning \citep{wong2023wordmodelsworldmodels} and decision-making abilities of LLMs \citep{conll-2023-babylm,park2023generative,wang2023voyageropenendedembodiedagent}. Games provide interactive and dynamic environments that can test models beyond static datasets. However, existing game-based benchmarks are often \emph{(i)} inflexible and limited in scope, \emph{(ii)} not easily extensible, \emph{(iii)} restricted in their effectiveness for comprehensive model evaluation, and \emph{(iv)} depend on predefined and hard-coded prompt templates.

To address these challenges, we introduce \ZSEVAL{}, a flexible and extensible open-source framework designed to evaluate LLMs \textit{dynamically} and \textit{relatively} through the simulation of games. Our framework allows for comprehensive assessment by providing models with multiple opportunities to make legal moves, thereby accommodating occasional errors and offering a more nuanced understanding of their capabilities.

Some important features of \ZSEVAL{} include:

\begin{enumerate}[wide, labelindent=0pt]
\item \textbf{Flexible and Extensible Framework}: \ZSEVAL{} is designed to be adaptable, allowing researchers and practitioners to customize and extend the evaluation environment to suit diverse needs.
\item \textbf{Robustness to Prompt Sensitivity}: By incorporating automatic prompt optimization, our framework mitigates issues related to prompt sensitivity, leading to more reliable evaluation outcomes.
\item \textbf{Enhanced Interpretability}: The structured environment and comprehensive logging facilitates easier interpretation of model behaviors, aiding in the identification of strengths and weaknesses.
\item \textbf{Error Accommodation}: Models are given multiple chances to make legal moves, ensuring that occasional missteps due to inherent stochasticity do not disproportionately affect the overall evaluation.
\end{enumerate}

\section{Related Work}
\label{sec:related_works}

\paragraph{Dynamic Evaluations}
To address the static benchmark issues highlighted in \Cref{sec:introduction}, the paradigm of evaluating agentic capabilities through simulations has been applied successfully in multiple prior works. Some notable ones include \textit{(i)} \textit{AgentBench} \citep{liu2024agentbench}, an evolving benchmark consisting of 8 environments that models interact with to complete tasks. \textit{(ii)} \textit{CRAB} \citep{xu2024crabcrossenvironmentagentbenchmark}, a benchmark for evaluating agentic behavior by executing tasks across multiple different environments. \textit{(iii)} KIEval \citep{yu2024kievalknowledgegroundedinteractiveevaluation}, a dynamic contamination-resilient evaluation framework: it engages the evaluated model in a dynamically generated and multi-turn conversation with another "interactor" model that attempts to extract whether a deep comprehension of the answer is present, or if it is solely memorized.
\paragraph{Game Evaluations}
There has been a substantial body of work on creating frameworks for evaluating LLMs on games. Some of these frameworks include ChatArena \citep{ChatArena}, GridGames \citep{topsakal2024evaluatinglargelanguagemodels}, GTBench \citep{duan2024gtbench}, SmartPlay \citep{wu2024smartplay}, and GameBench \citep{costarelli2024gamebenchevaluatingstrategicreasoning}. While the motivations of these works are closely similar to ours, they do not provide an easily extensible and general framework that allows for continuous evolution.
Furthermore, these works are specific to text-based game implementations. LVLM-Playground \cite{wang2025are} is a recent framework that was developed to test Large Vision Language Models on a variety of games that use both the language and vision modalities. While \ZSEVAL{} currently only has text-based games implemented, it also natively supports the implementation of multimodal games and player strategies, which is a promising direction discussed further in Section \ref{sec:future_work}.


\paragraph{Comparative Human Evaluations} A popular head-to-head LLM evaluation framework is Chatbot Arena\footnote{formerly LMSYS, not to be confused with ChatArena.} \citep{chatbotarena}, which allows users to prompt two anonymous LLMs with arbitrary prompts and to vote for the better response. This creates a diverse evaluation that effectively ranks all models in a leaderboard. However, it suffers from two issues: \textit{(i)} human evaluations are slow and laborious, and adding new models requires prolonged evaluation periods until sufficient votes are acquired for a confident placement, and \textit{(ii)} human evaluations contain human biases, such as prompt over-representation \citep{llama3arena2024} and bias to verbose and ``pretty'' responses \citep{chen2024humansllmsjudgestudy,park2024disentanglinglengthqualitydirect,stylearena2024}.

\section{Implementation}
The implementation of this framework closely follows the principle of completely separating game logic from player logic. Because of this, there are two axes which must be made easily extensible: adding games and adding player strategies. To ensure this, the respective classes are implemented in such a way that the developer only needs to know the logic of the game or strategy they are implementing. This minimizes the framework’s knowledge overhead and lowers the barrier to contribution.
\subsection{\texttt{GameState} Implementation}
\begin{figure}[ht]
    \centering
    \includegraphics[width=\linewidth]{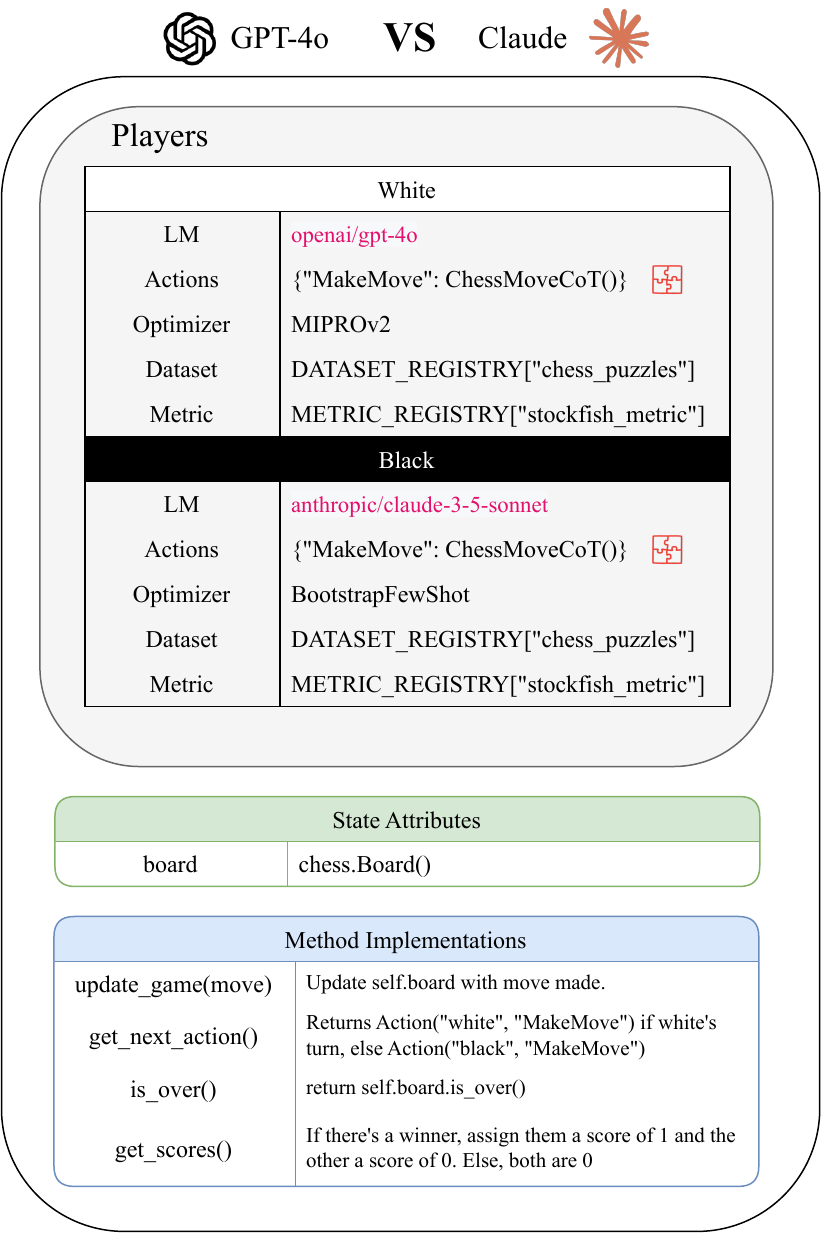}
    \caption{A high-level example implementation of the \texttt{GameState} class of Chess in \ZSEVAL{}.}
    \label{fig:chess_implementation}
\end{figure}

Drawing from extensive-form games in game theory \citep{osborne1994course} and Markov Decision Processes, we formalize a game in \ZSEVAL{} as the tuple
\begin{equation}
G = \langle S, U, P, A, R, \textit{Next}, \textit{Terminal} \rangle
\label{eq:game_state}
\end{equation}
where each component corresponds to a distinct concept in our framework (see Figure \ref{fig:chess_implementation} for an example implementation):

\begin{itemize}[wide, labelindent=0pt]
    \item \textbf{$S$ (State Space):} The set of all possible configurations of the game. In \ZSEVAL{}, this is represented by all the attributes of the \texttt{GameState} class (For example, the board state in the \texttt{ChessGame} class).
    
    \item \textbf{$U$ (Update/Transition Function):} A function that maps a state and the result of an action (a move) to a new state. This is implemented as \texttt{update\_game(move)}.
    
    \item \textbf{$P$ (Players):} The set of players participating in the game. These are defined by the \texttt{player\_definitions()} method and initialized in the \texttt{self.players} attribute of each game.
    
    \item \textbf{$A$ (Actions):} The set of possible actions available in the game. This is also specified in \texttt{player\_definitions()}, which returns a list of \texttt{PlayerDefinition} objects that detail each player's role and the actions it must implement.
    
    \item \textbf{$R$ (Reward/Score Function):} A function that maps a state to an assignment of scores (or rewards) for each player. This is provided by the \texttt{get\_scores()} method.
    
    \item \textit{Next} \textbf{(Next Action Function):} A function that maps a given state to a tuple containing the next action, the player responsible for that action, and the input $I$ provided to that player (which determines what each player observes). This functionality is implemented in the \texttt{get\_next\_action()} method.
    
    \item \textit{Terminal} \textbf{(Terminal/Over Condition):} A function that determines whether a state is terminal (i.e., the game has ended), mapping a state to a Boolean value (\texttt{true} or \texttt{false}). This is realized by the \texttt{is\_over()} method.
\end{itemize}

\subsection{\texttt{Player} Implementation}
Each game defines a set of player roles through player definitions. These definitions include the name of the player, the available actions they can take, and a default implementation when users do not specify their own.

Each player must have a clearly defined set of possible actions, with corresponding functions that determine how these actions are executed to generate moves.
What makes this framework particularly powerful is its integration with DSPy modules. Rather than implementing action functions directly, players can leverage DSPy modules to create sophisticated game-playing strategies that abstract away the complexities of prompt engineering.
\subsection{Why DSPy?}
DSPy modules offer a way to implement general game-playing strategies that abstract away prompting. This is beneficial for three main reasons:
\begin{enumerate}[wide, labelindent=0pt]
    \item \textbf{Higher-level Strategy Iteration:} Iterate on the level of programs rather than on the level of prompting. This allows for more complex strategies to be implemented and compared against each other. For example, a more complex DSPy program for a particular game could vastly outperform Chain-of-Thought prompting not because of the prompts themselves, but because of the logical structure of the program.
    \item \textbf{Prompt Sensitivity:} A strategy could perform very well on a particular model but not on another due to prompt selection that is less effective for certain models. By defining the pipeline using DSPy and optimizing for each model separately, this sensitivity would be minimized which would ensure that performance gains stem from the pipeline’s inherent logic rather than model-specific prompt tuning (assuming appropriate dataset and metric selection).
    \item \textbf{Native Retry Mechanism} 
    DSPy provides a structured way to handle errors and invalid model outputs through Assertions and Suggestions \citep{singhvi2024dspyassertionscomputationalconstraints}. By incorporating these assertions into the move-generation logic, the framework significantly reduces the number of "forgivable" failures, ensuring that a game continues smoothly unless the model consistently fails even after receiving feedback. This structured retry mechanism enhances game stability and minimizes disruptions caused by transient errors.
    \item \textbf{Ease of Module Sharing:} Optimized modules are easily saved and loaded which allows the community to compile and share modules of specific models performing well on specific games. This ability to share optimized modules allows for collaboration within the community which will accelerate research on the behavior of models in the games implemented in the framework.
\end{enumerate}

\subsection{Streamlining Prompt Optimization}
\ZSEVAL{} streamlines prompt optimization by automating the process within each class extending \texttt{Player}, and by creating a registry system for datasets and metrics. Developers need only to register their dataset and metric and specify the optimization configuration when initializing a player, which further reduces the need for boilerplate code and accelerates development. The optimized modules are automatically cached based on the optimizer, dataset, and metric configurations.

\subsection{Game Management}
\begin{figure}
    \centering
    \includegraphics[width=\linewidth]{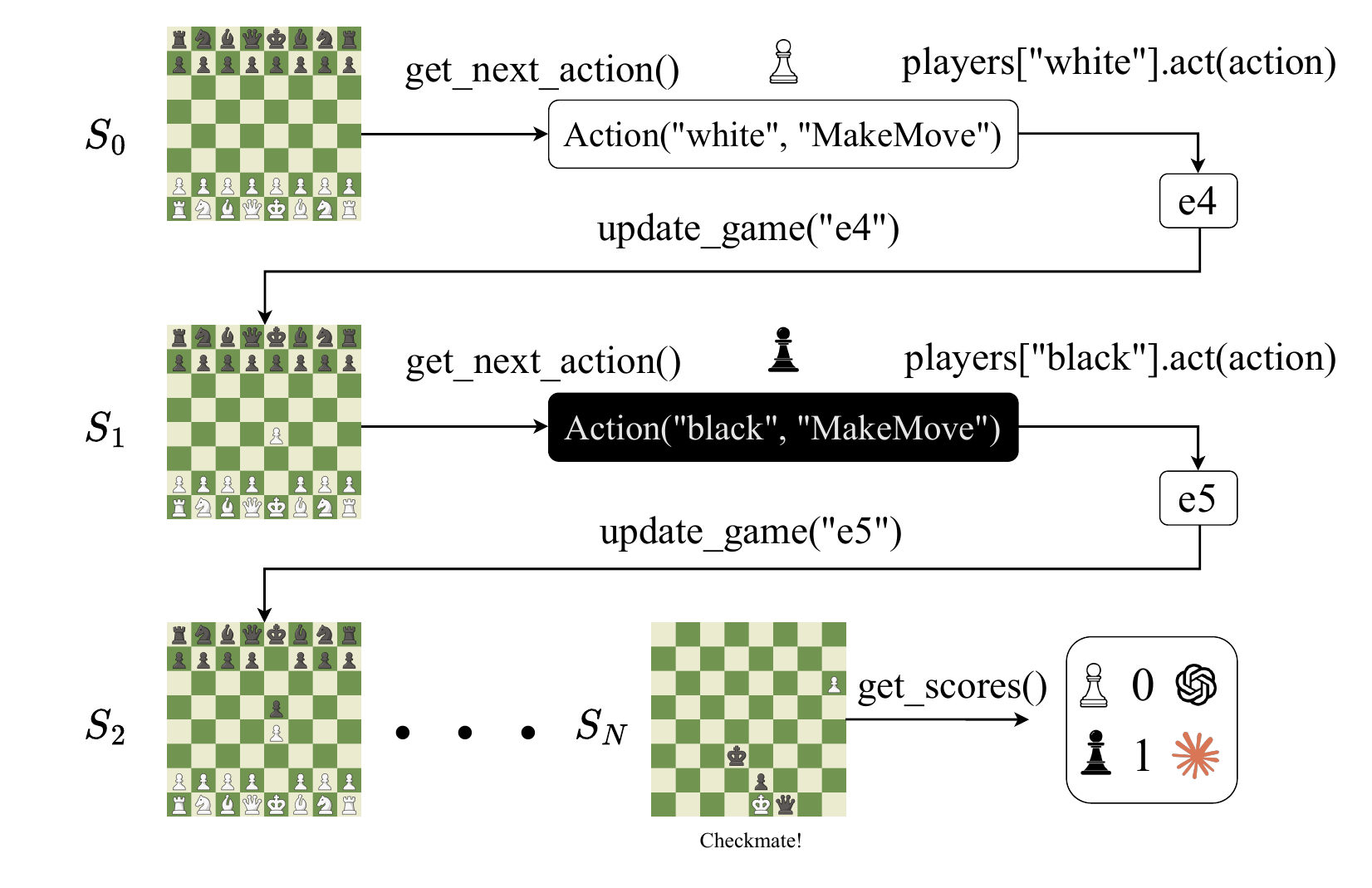}
    \caption{An example flow of the Game Manager for the game of Chess. The state of the game moves forward by \textit{(i)} querying the current state for the next action and the player that is expected to act \textit{(ii)} executing that action using the player's implementation for that action, \textit{(iii)} updating the game state with that action, \textit{(iv)} repeat i-iii until the game is terminated. The scores are then calculated from the final state and a winner is determined accordingly.}
    \label{fig:chess_manager}
\end{figure}

\ZSEVAL{} also implements game management classes that ease the running of games. The \texttt{GameManager} class handles the running of a single game (see Figure \ref{fig:chess_manager}). \texttt{GamePoolManager} uses this class to extend it to run a "pool" of games between any number of specified models. It matches language models against each other given a matching strategy like Round-Robin and keeps count of the wins, draws, and losses of each model.

\subsection{Rating}
Following recent suggestions for head-to-head LLM rating systems by \citet{boubdir2023elouncoveredrobustnessbest,eloarena2023}, we employ the \citet{bradleyterryrank} (BT) rating system, an alternative to the \citet{elo} system, to rate models. The BT model is permutation-invariant and assumes a fixed win rate for each pair of models, maximizing the likelihood of observed outcomes. This choice is more suitable than the traditional Elo system, which was designed for human chess players with varying skill levels, whereas LLMs have fixed skill levels defined by their weights.

\section{Example Games}
\ZSEVAL{} currently supports a total of 7 games:

\begin{itemize}[wide, labelindent=0pt]
    \item \textbf{Debate}: Given a topic, players start by giving opening statements then take turns giving rebuttals before a jury of LLMs scores each side based on a well-defined numerical rubric to minimize LLM-as-a-judge bias.
    \item \textbf{Chess}: The game of chess implemented such that players have multiple chances in making a valid move both in format (FEN) and in game rules.
    \item \textbf{Poker}: A simple variant of Texas Hold 'Em that allows up to 10 players.
    \item \textbf{Gandalf}: Directly inspired from the game with the same name\footnote{\url{https://gandalf.lakera.ai}}, this game assigns one player the role of the Sentinel, where their objective is to make conversation without revealing a secret password to the Infiltrator.
    \item \textbf{Liar's Dice}: A simple bluffing game where players take turns bidding on dice or calling the other player's bluff.
    \item \textbf{MathQuiz}: An adversarial game where one player with the Teacher role generates a difficult math question that it can solve itself but not the other player with the role of Student.
    \item \textbf{PyJail}: A CTF-like challenge where one player writes a \texttt{jail(user\_input)} function. The other player is then given a number of attempts to try different inputs and observe the output with the goal of getting access to the flag stored in an environment variable.
\end{itemize}

These games cover a wide variety of capabilities such as reasoning (Chess, Poker), conversational skills (Gandalf), argumentation (Debate), and security (PyJail).

\paragraph{Scalable Verification}
The MathQuiz and PyJail games require competing models to generate complex challenge environments and solutions. Since verification of the knowledge-based challenges by a human in the loop is not scalable, we design a method to verify model output using an automated manager in a two-step generation and verification process. This is accomplished by defining a target outcome (e.g., the answer to a math question or a CTF flag) as the basis for verifying generated input, and regulating the model context at each stage.

The exact process (illustrated in \Cref{fig:state-diagram-verification}) is outlined as follows:
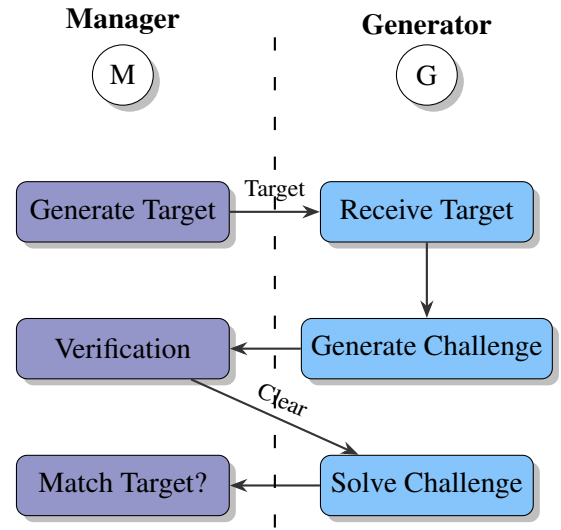
\begin{figure}
    \centering
    \begin{tikzpicture}[
        box/.style={rectangle, draw, rounded corners, fill=zse_color1!50, drop shadow, minimum height=0.8cm, minimum width=2.8cm, align=center},
        boxllm/.style={rectangle, draw, rounded corners, fill=zse_color2!50, drop shadow, minimum height=0.8cm, minimum width=2.8cm, align=center},
        actor/.style={circle, fill=white!10, draw, drop shadow, minimum size=0.8cm},
        arrow/.style={-Stealth, thick, draw=black!80},
        node distance=1cm and 2cm
    ]
    \node[actor, label=above:{\textbf{Manager}}] (manager) at (-2, 0.5) {M};
    \node[actor, label=above:{\textbf{Generator}}] (generator) at (2, 0.5) {G};
    
    \node[box, below=of manager] (generate) {Generate Target};
    \node[box, below=of generate] (verify) {Verification};
    \node[box, below=of verify] (match) {Match Target?};
    
    \node[boxllm, below=of generator] (genModel) {Receive Target};
    \node[boxllm, below=of genModel] (output) {Generate Challenge};
    \node[boxllm, below=of output] (clear) {Solve Challenge};
    
    \draw[arrow] (generate) -- (genModel) node[midway, above, font=\small] {Target};
    \draw[arrow] (genModel) -- (output);
    \draw[arrow] (output) -- (verify);
    \draw[arrow] (verify) -- (clear) node[midway, above, sloped, font=\small] {Clear};
    \draw[arrow] (clear) -- (match);
    
    \draw[dashed, thick, dash pattern=on 5pt off 10pt] (0, 1) -- (0, -5.5);
    \end{tikzpicture}
    \caption{State diagram of the verification process involving the Game Manager and the Generator. Purple boxes indicate deterministic steps and blue boxes indicate steps involving the model.}
    \label{fig:state-diagram-verification}
\end{figure}
\begin{enumerate}[wide, labelindent=0pt]
\item The generator model receives a target and attempts to output a valid challenge that resolves to the specific target.

\item In the verification step, the manager restricts the model's context to ensure no direct access to the target, and asks the generator model to solve the previously generated challenge.

\item If the manager determines the verification is successful (by matching the target with the generator's solution), the game proceeds. Otherwise, the generator model is deemed to have failed to generate a valid challenge.
\end{enumerate}
This method ensures the generated challenge environment is valid and a solution is proven possible by the generator. The design also correctly penalizes models that directly generate memorized questions as it is likely to have been memorized by other models, thereby encouraging models to create challenging and novel questions. Finally, the scalability of the evaluation is preserved as the capabilities of models scale.


\section{Results}

\begin{figure}
    \centering
    \includegraphics[width=0.9\linewidth]{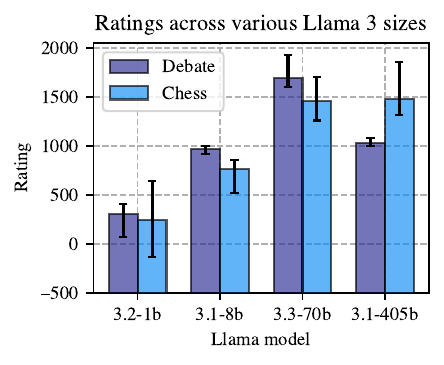}
    \caption{Ratings of Llama 3 models of various subversions and sizes placed head-to-head. error bars are 95\% confidence intervals of BT ratings obtained via bootstrapping.}
    \label{fig:llama_ratings}
\end{figure}

\Cref{fig:llama_ratings} shows the outcome of placing various Llama 3 \citep{llama3} models head-to-head in two games: chess and debate. As expected, there is a clear positive correlation between model size and performance in both games, with the only exception being that Llama 3.3 70B outperforms Llama 3.1 405B in debate, this is likely due to the more refined fine-tuning approach taken in the 3.3 version compared to 3.1\footnote{\url{https://github.com/meta-llama/llama-models/blob/main/models/llama3_3/MODEL_CARD.md}}. We expect to observe such interesting results as the use of \ZSEVAL{} expands with more tested models and implemented games.

\section{Future Work}
\label{sec:future_work}
This work serves as a launchpad for researchers and practitioners to further explore the paradigm of LLM evaluation through competition. One key avenue for investigation is the impact of prompt optimization on final rankings. Previous research has shown that leaderboards can be highly sensitive to minor perturbations in benchmarks \citep{alzahrani2024benchmarks}. Could prompt optimization help stabilize rankings and mitigate these instabilities? Additionally, how might one go about setting up a leaderboard using \ZSEVAL{}?

Another promising direction is the integration of games requiring multi-modal capabilities. While the current implementation focuses on text-based games, \ZSEVAL{} is designed to support any type of game. For instance, in a board game setting, instead of representing the game state as a string—which can be convoluted for certain games like Diplomacy—an image-based representation could convey the same information more efficiently. This concept could be extended further to include full 3D simulations, where models process rendered environments as input. Recent work has demonstrated the efficacy of this direction on Large Vision Language Models \citep{wang2025are}.

The competitive evaluation paradigm also lends itself naturally to adversarial strategies, making it particularly well-suited for assessing models in security-focused games. As an initial step in this direction, we implemented PyJail as a simple example, but we envision much more sophisticated environments that could push this approach even further.

\section{Conclusion}
    The dynamic, relative, and competitive nature of the \ZSEVAL{} framework lays the groundwork for a more robust and trustworthy measurement of AI model capabilities, advancing the state of benchmarking in LLMs. By leveraging games, we ensure that models are consistently challenged with diverse, evolving tasks, minimizing the risk of overfitting and saturation commonly observed in static benchmarks. Additionally, the close integration of DSPy provides an abstraction layer that allows for easily implementing and testing different strategies, easily retrying, and reduced prompt sensitivity owing to DSPy's collection of prompt optimization algorithms.

\bibliography{citations}


\end{document}